\documentclass{ieeeaccess}

\usepackage{soul}
\usepackage{algorithm}
\usepackage{algorithmic}
\usepackage{listings}
\usepackage{multirow}
\usepackage{lineno}
\usepackage{hyperref}
\usepackage{booktabs}
\usepackage{mathtools}
\soulregister\cite7
\soulregister\ref7
\soulregister\sc7
\usepackage{eurosym}
\usepackage{siunitx}
\usepackage{textgreek}
\usepackage{tablefootnote}
\renewcommand{\theenumi}{\Alph{enumi}}
\usepackage[caption=false,font=normalsize,labelfont=sf,textfont=sf]{subfig}

\def\textsc#1{\textnormal{{\sc #1}}}
\usepackage{caption}
\DeclareCaptionFont{ieeeblue}{\color{accessblue}}
\DeclareCaptionLabelFormat{myformat}{\raggedright\figcapfont{\textbf{#1}\textbf{#2}}}
\def\BibTeX{{\rm B\kern-.05em{\sc i\kern-.025em b}\kern-.08em
T\kern-.1667em\lower.7ex\hbox{E}\kern-.125emX}}

\newcommand{\expnumber}[2]{{#1}\text{e}\textsuperscript{#2}}
\newcommand{\sctxt}[1]{{\sc #1}}
\newcommand{\red}[1]{{\color{red} #1}}
\soulregister{\sctxt}{7}
\soulregister{\red}{7}
\soulregister{\expnumber}{7}
\soulregister{\SI}{7}
\soulregister{\percent}{7}

\usepackage{blindtext}
\usepackage{hyperref}
\usepackage{nameref}

\newcounter{mylabelcounter}

\makeatletter
\newcommand{\labelText}[2]{%
\refstepcounter{mylabelcounter}%
\immediate\write\@auxout{%
 \string\newlabel{#2}{{\unexpanded{#1}}{\thepage}{{\unexpanded{#1}}}{mylabelcounter.\number\value{mylabelcounter}}{}}%
}
}
\makeatother

\makeatletter
\newcommand\footnoteref[1]{\protected@xdef\@thefnmark{\ref{#1}}\@footnotemark}
\makeatother

\newcommand{\olsi}[1]{\,\overline{\!{#1}}}
\newcommand{\sfunction}[1]{\texttt{\textbf{#1}}}

\begin{document}

\history{Date of publication 28 June 2024.}
\doi{10.1109/ACCESS.2024.3420710}

\title{Predictability and Causality in Spanish and English Natural Language Generation}

\author{\uppercase{Andrea Busto-Castiñeira}\authorrefmark{1},
\uppercase{Francisco J. González-Castaño}\authorrefmark{1}, \uppercase{Silvia García-Méndez}\authorrefmark{1}, and \uppercase{Francisco de Arriba-Pérez}\authorrefmark{1}.}

\address[1]{Information Technologies Group, atlanTTic, Telecommunication Engineering School, University of Vigo, 36310 Vigo, Spain}

\tfootnote{This work was supported in part by Xunta de Galicia grants ED481B-2021-118, ED481B-2022-093, and ED431C 2022/04, Spain; Ministerio de Educación grant FPU21/00798, Spain; Ministerio de Ciencia e Innovación grant TED2021-130824B-C21, Spain; and by the Galician Supercomputing Center (CESGA), Spain, which provided the computing resources required.}

\markboth
{A. Busto-Castiñeira \headeretal: Analytical Study of Predictability and Causality in
Spanish and English Natural Language Generation}
{A. Busto-Castiñeira \headeretal: Analytical Study of Predictability and Causality in
Spanish and English Natural Language Generation}

\corresp{Corresponding author: Andrea Busto-Castiñeira (e-mail: abusto@gti.uvigo.es).}

\begin{abstract}
In recent years, the field of Natural Language Generation (NLG) has been boosted by the recent advances in deep learning technologies. Nonetheless, these new data-intensive methods introduce language-dependent disparities in NLG as the main training data sets are in English. Also, most neural NLG systems use decoder-only (causal) transformer language models, which work well for English, but were not designed with other languages in mind. In this work we depart from the hypothesis that they may introduce generation bias in 
target languages with less rigid word ordering, subject omission, or different attachment preferences for relative clauses, so that for these target languages other language generation strategies may be more desirable. 

This paper first compares causal and non-causal language modeling for English and Spanish, two languages with different grammatical structures and over 1.5 billion and 0.5 billion speakers, respectively. For this purpose, we define a novel metric of average causal and non-causal context-conditioned entropy of the grammatical category distribution for both languages as an information-theoretic a priori approach. 

The evaluation of natural text sources (such as training data) in both languages reveals lower average non-causal conditional entropy in Spanish and lower causal conditional entropy in English. According to this experiment, Spanish is more predictable than English given a non-causal context. Then, by applying a conditional relative entropy metric to text generation experiments, we obtain as insights that the best performance is respectively achieved with causal NLG in English, and with non-causal NLG in Spanish. These insights support further research in NLG in Spanish using bidirectional transformer language models.
\end{abstract}


\titlepgskip=-21pt

\maketitle

\section{Introduction}
\label{sec:introduction}
\PARstart{T}{hanks}  to their capacity to acquire universal language representations from vast amounts of unlabeled text data, transformer-based Natural Language Generation (NLG) models \cite{attention-2017} have achieved unprecedented success \cite{Lin2022}. 

Transformers are based on Sequence-to-Sequence (seq2seq) language models \cite{seq2seq-2014} by enhancing them with positional encoding, which enables parallel training while still considering word order, and a novel self-attention mechanism that selects the most relevant parts of the input sequence. The unsupervised nature of transformers' pre-training facilitates handling vast raw text data.

However, the linguistic prevalence of English on the Internet is a primary source of bias \cite{Garrido-Munoz2021}. Most evaluation data sets and benchmarks are primarily or entirely written in English with very few exceptions \cite{Clark2020}, and therefore the most innovative contributions rarely target other languages. This linguistic data imbalance has a detrimental effect on polyglot language models' word embeddings \cite{Liu2020, Xue2021}, as tokenizers assign longer tokens to character sequences in languages with more extensive representation \cite{Wu2020a, Rust2021, Erdem2022}, despite of some proposed solutions such as those in \cite{Artetxe2019} and \cite{Clark2022}.
The multilingual NLP paradigm of cross-lingual transfer learning, which tends to view non-English languages as particular use cases, makes language models favor English-like grammars with more strict word ordering and explicit subject \cite{Guarasci2022}.

Nowadays, the majority of generative language models are decoder-only transformers, being OpenAI's GPT-3/3.5/4 \cite{gpt3-2020}, and ChatGPT the most popular, and Big Science's BLOOM, Meta AI's OPT, and Google AI's BARD also well-known examples. Most of these models are multilingual, however, their performance varies between languages. This led to monolingual implementations of the smaller model GPT-2 in languages other than English, for example in German\footnote{\label{cedille}Available at: \texttt{\url{https://cedille.ai}}, June 2024.}, French\footref{cedille}, Italian \cite{demattei2020geppetto}, and Spanish \cite{gutierrezfandino2022}.

These generative models are exclusively \textit{causal}, that is, they produce text from left to right by recursively feeding the model with previously generated sequences. 
As in the case of Recurrent Neural Networks (RNN), decoder-only transformers are expectation-based word predictors. These systems tend to favor structures in which related elements are close along the sequence, such as relative clause attachments to syntactically lower nominals in ambiguous contexts, which fits nicely into English syntax \cite{Davis2020}.

However, the mutually beneficial congruence between causal language modeling and English may not apply to other languages. Not only does Spanish prefer a higher nominal attachment in the resolution of ambiguous relative clauses, but its syntax is also highly flexible, even within declarative sentences \cite{Lahousse2012}. This is strongly opposed to the more strict subject-verb-object structure of the English language, which allows for few inversion exceptions \cite{Assaiqeli2021}.

Unlike causal language models, encoder-only \textit{non-causal} language models generate word embeddings using bidirectional contexts, which means that the model output can be conditioned by both left and right tokens. This eliminates the output sequence's sequential dependencies and allows alternative generation orders.

In light of this, we depart from the hypothesis that decoder-only (causal) transformer language models may introduce generation bias in target languages with less rigid word ordering than English, subject omission, or different attachment preferences for relative clauses, so that for these target languages other language generation strategies may be more desirable.  To put this hypothesis to test, in addition to English, we consider and Spanish, a language with a different grammatical structure and also a broad base of speakers (these languages sum over 1.5 billion and 0.5 billion speakers, respectively, a substantial share of the world's population).  However, the approaches in this work can be extended to obtain insights on other languages and NLP tasks. 

Our contributions are:

\begin{enumerate}
\item First, we present a novel information-theoretic approach to study language predictability. We compare the causal context-conditioned entropy and the non-causal context-conditioned entropy of the grammatical category distribution of source natural texts to assess whether their language is more predictable from causal or non-causal language contexts. This reveals lower average non-causal conditional entropy in Spanish  and lower causal conditional entropy in English. According to this assessment, Spanish is more predictable than English given a non-causal context.

\item Then, using both automatic (based on conditional relative entropy) and manual evaluation methodologies, we put decoder-only and encoder-only transformer language models to test to assess empirical causal and non-causal NLG performance, seeking to evaluate if the currently dominant causal NLG paradigm is adequate from a language-agnostic perspective or whether specific languages may benefit from other word generation orderings. We obtain as insights that the best performance is achieved with causal NLG in English and non-causal NLG in Spanish. These insights support further research in NLG in Spanish using bidirectional transformer language models instead of the dominant decoder-only ones.
\end{enumerate}

The rest of this paper is organized as follows. Section \ref{sec:related} reviews related work on both psycholinguistic language predictability and language model causality in NLG. Section \ref{sec:methodology} describes the proposed analytical methodology used for the experiments. Sections \ref{sec:predictability-results} and \ref{sec:text-generation-results} present the details and results of the assessments of predictability and text generation performance, respectively. Section \ref{sec:discussion} summarizes and discusses the results obtained. Finally, Section \ref{sec:conclusion} concludes the paper.

\section{Related work}
\label{sec:related}

In this section, we discuss relevant works on both causality in NLG (Section \ref{sec:noncausal_related}) and language predictability (Section \ref{sec:predictability}).

\subsection{Causality in generative transformer language models}
\label{sec:noncausal_related}
The contextual awareness of a transformer is controlled by self-attention. The base concept behind this attention mechanism is a mapping of a query ($\mathbf{q}$) into pairs of keys ($\mathbf{k}$) and values ($\mathbf{v}$). By respectively denoting the queries', keys', and value sets' matrices as $\mathbf{Q}$, $\mathbf{K}$ and $\mathbf{V}$, we define self-attention as:

\begin{equation}
  Attention\left(\mathbf{Q}, \mathbf{K}, \mathbf{V}\right) = softmax\left(\frac{\mathbf{Q} \mathbf{K}^{T}}{\sqrt{\left|\mathbf{k}\right|}}\right)\mathbf{V}
\end{equation}

Transformers, rather than a single attention function, project queries, keys, and values onto $h$ separate heads. This is called multi-head attention:

\begin{equation}
    MultiHead\left(\mathbf{Q}, \mathbf{K}, \mathbf{V}\right) = concat\left(\mathrm{head}_1, \cdots, \mathrm{head}_h \right)\mathbf{W}^{O}
\end{equation}

By denoting each head attention function as:

\begin{equation}
    \mathrm{head}_i = Attention\left(\mathbf{Q} \mathbf{W}_i^Q, \mathbf{K} \mathbf{W}_i^K, \mathbf{V} \mathbf{W}_i^V\right)
\end{equation}

 where $\mathbf W_i^Q$, $\mathbf W_i^K$, $\mathbf W_i^V$ and $\mathbf W^{O}$ are parameter projection matrices for the queries, keys, values, and output respectively.

This attention mechanism is present in all the layers of both the encoder and the decoder, if present. While the encoder's attention is bidirectional, the decoder has two different types of attention: (\textit{i}) a masked multi-head attention block that masks non-causal context and (\textit{ii}) a bidirectional multi-head attention block that receives non-causal information from the encoder.

Even though this encoder-decoder architecture is popular in some NLP tasks such as machine translation \cite{Chen2020, Wu2020b, Kawara2021, Nguyen2021}, several transformer-based models only have one of these components. By omitting the encoder in decoder-only transformers, all non-causal contextual dependencies are removed by exclusively using masked attention. Decoder-only transformers are nowadays the best performing task-agnostic NLG systems. Nevertheless, there exist some state-of-the-art non-causal NLG solutions. For example, non-causal language models can be trained for the Masked Language Modeling (MLM) objective, a task in which the language model predicts masked words within a sentence \cite{Zeng2021}. Typically, non-causal NLG systems are focused on particular tasks such as speech recognition \cite{Bai2021, Chen2021, Wang2022}, style transfer and grammar correction \cite{Kaneko2020}, textual data augmentation \cite{Park2019}, and task-specific dialog systems \cite{Balaraman2021, Yu2021}. 

\subsection{Language predictability}
\label{sec:predictability}
Conditional entropy is a typical metric for evaluating the predictability of a problem given its input variables and expected output probability distribution \cite{Song2010,Li2022}. Conditional entropy $H(X\mid Y)$ measures the extra information carried by a variable $X$ when another conditional variable $Y$ is available as side information.

In psycholinguistics, surprisal theory also uses this information-theoretic concept to quantify processing difficulty in sentence comprehension \cite{Hale2001, Levy2008}.
Multiple studies provide empirical evidence for this expectation-based theory by showing correlations between textual surprisal and both test subjects' reading times, as in \cite{Lowder2018}, and brain activity, as in \cite{Henderson2016}.

Even if generally accepted, surprisal theory does not model working memory in text comprehension, disregarding processing difficulties in integrating words or components that are widely apart within a text \cite{Gibson1998, Lewis2005, Bartek2011, Nicenboim2015, Nicenboim2016}. Lossy context surprisal \cite{Futrell2020} combines expectation and memory-based predictability theories by modeling working memory constraints as noise. Even though this model premise is independent of language, it can accurately reflect several language-specific text-processing phenomena. 

Lossy context surprisal recreates structural forgetting by dropping part of the context and re-sampling it incorrectly from the a priori language knowledge probability model. Structural forgetting \cite{Vasishth2010} is a common grammatical illusion in English in which ungrammatical double-embedded relative clauses can be perceived as correct. 
Probabilistic language expectations can determine this exclusively. In \cite{Frank2016} it is proven that native and non-native speakers show structural forgetting in English, but do not behave this way when presented with the same syntactic structures in German or Dutch. This propensity of English probabilistic distribution to such backward prediction mistakes is coherent with the issue of non-causal English text generation. 

The main goals of neural NLP and psycholinguistics approaches to language cognition are very similar: (\textit{i}) to give formally explicit descriptions of the mental structures underpinning cognitive processes, and (\textit{ii}) to explain the learning mechanisms behind them \cite{Pater2019}. Even if research in these areas tends to diverge, recent contributions to the study of linguistic theory use language models \cite{Hofmann2022}, further evidencing their alignment. 

However, to our knowledge, psycholinguistics concepts have yet to be applied to neural language modeling other than for data set elaboration \cite{Linzen2019}. With this in mind, our work provides a novel linguistic-based conditional entropy hypothesis test for language modeling causality (see contribution 1 above), whose findings can support future NLG designs and methodologies.

\section{Methodology}
\label{sec:methodology}

\subsection{Predictability hypothesis test}
\label{sec:methodology-hypothesis}
Causal language models predict the next token in a sequence of tokens. These models are solely concerned with the left context for sinistrodextral (\textit{i.e.}, written from left to right) languages such as Spanish and English (conversely, non-causal models trained on the MLM task consider the bidirectional context for blank-filling-based text generation).

Given a sequence of tokens $X$ as context, language models provide the probability mass function for the next predicted token $\hat{X}$. For a generation index $i<N$, we define the $N$-long input causal context as follows:
\begin{equation}
    \mathbf{x}_{c_i} = \begin{bmatrix}x_{i-N} & \dots & x_{i-1} \end{bmatrix}^T
\end{equation}

And the non-causal context as:

\begin{equation}
\mathbf{x}_{n_i}  = \begin{bmatrix} x_0 & \dots & x_{i-1} & x_{i+1} & \dots & x_n \end{bmatrix}^T
\end{equation}

So that we can express the output of a causal language model as:

\begin{equation}
    \mathbf{y}_{c_i} = \begin{bmatrix} p(\hat{X}_i = v_0 \mid  X_c = \mathbf{x}_{c_i}) \\ \vdots \\ p(\hat{X}_i = v_{\left|\mathcal{V}\right|-1} \mid  X_c = \mathbf{x}_{c_i} ) \end{bmatrix}
\end{equation}

And the output of a non-causal language model as follows:

\begin{equation}
    \mathbf{y}_{n_i} = \begin{bmatrix} p(\hat{X}_i = v_0 \mid  X_n = \mathbf{x}_{n_i}) \\ \vdots \\ p(\hat{X}_i = v_{\left|\mathcal{V}\right|-1} \mid  X_n = \mathbf{x}_{n_i} ) \end{bmatrix}
\end{equation}

with vocabulary set $\mathcal{V} = \left\{\begin{matrix} v_0 & \dots & v_{\left|\mathcal{V}\right|-1}\end{matrix}\right\}$ of size $\left|\mathcal{V}\right|$.

As stated in Section \ref{sec:predictability}, we use a novel metric of conditional entropy to test whether a language is more or less predictable given causal or non-causal contexts (and thus, for example, whether Spanish NLG may benefit from non-causal language generation ordering). The less conditional entropy a problem has, the more predictable its outcome. As NLG is a language prediction task in which previously generated words are available as context, we want to compare the conditional entropy in two scenarios: (\textit{i}) causal text generation, in which text is generated from left to right so that we provide words to the left of the predicted one as context; and (\textit{ii}) non-causal text generation, which uses both left and right context for word prediction.

In order to test the predictability of causal and non-causal language models for English and Spanish, we compute and compare the average causal and non-causal conditional entropy for textual data in both languages:

\begin{equation}
    \label{eq:ent_causal}
    \olsi{H(\hat{X}\mid X_c)} = \sum_{\mathbf{x}_c \in \mathcal{X}^N_c} p(X_c = \mathbf{x}_c) H(\hat{X}\mid X_c = \mathbf{x}_c)
\end{equation}

\begin{equation}
    \label{eq:ent_non-causal}
    \olsi{H(\hat{X}\mid X_n)} = \sum_{\mathbf{x}_n \in \mathcal{X}^N_n} p(X_n = \mathbf{x}_n) H(\hat{X}\mid X_n = \mathbf{x}_n)
\end{equation}

with:

\begin{equation}
H(\hat{X}\mid X = \mathbf{x}) =
 \sum_{\hat{x}\in\mathcal{V}} p(\hat{x}\mid X = \mathbf{x}) \log \frac{1}{p( \hat{x}\mid X=\mathbf{x})}
\end{equation} 

\vspace{10pt}

It must be noted that both $\mathcal{X}^N_c$ and $\mathcal{X}^N_n$ have size $\left|\mathcal{V}\right|^N$. Context length and vocabulary size determine the accuracy of our estimation results. 
As we have no previous information about token probability distribution, we model both $p(X)$ and $p(\hat{X} = \hat{x}\mid X = \mathbf{x})$ as categorical distributions. The estimators used for these distributions are:

\begin{equation}
    \widehat{Pr}(X=\mathbf{x}_i) = \frac{L_i}{L}
\end{equation}

and

\begin{equation}
\widehat{Pr}(\hat{X} = x_j \mid X = \mathbf{x}_i) = \frac{L_{i_j}}{L_i}
\end{equation}

with $L$ being the token sequence length, $L_i$ the number of instances of the context $i$, and $L_{i_j}$ the number of instances in which token $j$ appears given context $i$.

In case both $p(X)$ and $p(\hat{X} = \hat{x}\mid X = \mathbf{x})$ are discrete uniform distributions, these estimators have normalized variances $\frac{\left|\mathcal{V}\right|^N - 1}{L}$ and $\frac{\left|\mathcal{V}\right|^N\left(\left|\mathcal{V}\right| - 1\right)}{L}$, respectively. This means that, in order to set our estimators' normalized variances to a specific value, the number of analyzed tokens should be proportional to $\left|\mathcal{V}\right|^N$ and $\left|\mathcal{V}\right|^{N+1}$, respectively.

Therefore, given the data available, neither word nor subword tokenization are feasible. We instead use a grammatical categorization based on Part-Of-Speech (POS) tagging. It reduces vocabulary size and data requirements dramatically while maintaining the original goal.

The resulting hypothesis test evaluates how predictable natural English and Spanish syntaxes are for causal and non-causal language models. Our first intuition is that non-causal predictability, as determined by the inverse of the non-causal context-conditioned entropy, will be higher for Spanish than for English and the opposite for causal predictability. By validating this, we can demonstrate that causal ordering may not be ideal for Spanish NLG, paving the way for further study of non-causal Spanish text generation approaches based on bidirectional transformers.

\subsection{Non-causal text generation}
For non-causal NLG, first we start with a sequence of [MASK] tokens of the desired length $K$. At each iteration, we re-sample every token once. We mask and fill tokens in groups of size $N$. In order to fill the masked tokens, we sample the output of a non-causal language model, from which we remove adjacent tokens, short prefixes and suffixes, and unknown tokens to enhance the overall quality of the produced sequence. This process is formally described in Algorithm \ref{al:nc-gen}.

\begin{algorithm}[bt]
\caption{Non-causal text generation.}\label{al:nc-gen}
\begin{algorithmic}
\STATE $\mathbf{x}  \gets \begin{bmatrix} \mathrm{[MASK]} & \dots & \mathrm{[MASK]}\end{bmatrix}_{K}$
\STATE $i \gets 0$
\WHILE {$i$ $<$ $I$}
    \STATE \textit{\textbf{index}} $\gets \sfunction{Shuffle}  \left(\begin{bmatrix} 0 & \cdots && K-1\end{bmatrix}\right)$
    \STATE $i \gets 0$
    \WHILE {$j$ $<$ $K - N$}
        \STATE \textit{\textbf{masked}} $\gets$ \textit{\textbf{index}}$\left[\begin{bmatrix} j & \cdots & \min\left(j+  N, K-1\right)\end{bmatrix}\right]$
        \STATE $x[$\textit{\textbf{masked}}$] \gets$ [MASK]
        \STATE $\mathbf{y} \gets \sfunction{NonCausalLM}\left(\mathbf{x}\right)$
        \FORALL{$m\in$ \textit{\textbf{masked}}}
            \STATE \textit{\textbf{prob}, \textbf{idx}} $\gets \sfunction{Filter}\left(\mathbf{y} \left[m\right]\right)$
            \STATE $\mathbf{x} \left[m\right] \gets \sfunction{Sample}$(\textit{\textbf{prob}, \textbf{idx}})  
        \ENDFOR
        \STATE $j \gets j+N$
    \ENDWHILE
    \STATE $i ++$
\ENDWHILE
\end{algorithmic}
\end{algorithm}

The number of iterations $I$ and the number of tokens masked in each generation step $N$ must be predetermined. These parameters influence the performance and computational efficiency of the algorithm. More masked tokens per generation step mean fewer calls to the language model function ($\lceil\frac{K}{N}\rceil$ calls per iteration), resulting in improved computing efficiency. In this work we set $N=2$ and $I=30$.

\subsection{Automatic evaluation}
\label{sec:auto_metrics}
The relative entropy $\mathcal{D}_{KL}\left(P\mid\mid Q\right)$, also known as KL divergence, quantifies the expected increase in uncertainty that comes from modeling a reference distribution $P$ as another distribution $Q$.
In this study, we use the following formulation for the conditional relative entropy for both causal ($X=X_c$) and non-causal ($X=X_n$) contexts:

\begin{equation}
\label{eq:eval_metric}
\begin{split}
     \mathcal{D}_{KL}\left(P(\hat{X}\mid X)\mid\mid Q(\hat{X}\mid X)\right) =& \sum_{\mathbf{x} \in \mathcal{X}^N} q(\mathbf{x}) \sum_{\hat{x} \in \mathcal{V}}p(\hat{x}\mid X = \mathbf{x})\\
     & \log \frac{p(\hat{x}\mid X = \mathbf{x})}{q(\hat{x}\mid X = \mathbf{x})}
\end{split}
\end{equation}

where $p$ and $q$ are the conditional probability density functions of our reference textual dataset's POS tags and the sequences to evaluate, respectively.

\subsection{Manual evaluation}
\label{sec:human_method}
The annotators were asked yes/no questions  on the following aspects to assess the quality of the generated sequences:

\begin{itemize}
    \item \textbf{Q1. Concordance}, penalizing improper use of verb tenses, number, and, if applicable, gender of determinants, adjectives, nouns, and pronouns.
    \item \textbf{Q2. Syntactic structure correctness}, by checking if all sequences have at least one subject and one verb and assessing that the sentences are syntactically sound in general. 
    \item \textbf{Q3. Word or phrase-level repetitions}, by penalizing word redundancy, duplication in enumerations, and subject redundancy, while trying to respect those repetitions that may be considered stylistic choices.
    \item \textbf{Q4. Word sense}. Language models can generate new words by combining prefixes, suffixes, and pronouns as sequential tokens. This question penalizes nonsensical words.
\end{itemize} 

We assessed inter-agreement with accuracy and $\alpha$-reliability \cite{Krippendorff2012} 
to verify that the annotations were neither arbitrary ($acc = \alpha = 0$) nor redundant ($acc = \alpha = 1$).

Finally, we also included a more general rating question (Q5) in which we asked annotators to provide a numerical score between 1 and 5 based on their impression of the annotation and their experience.

\section{Predictability test results}
\label{sec:predictability-results}

We used two different experimental setups to perform the hypothesis test in Section \ref{sec:methodology-hypothesis}: (\textit{i}) A first setup using two data sets, one in English and another in Spanish, which are exact translations of each other (Section \ref{sec:tale_data}); and (\textit{ii}) another setup with larger, relatively similar English and Spanish data sets that cannot be considered exact translations of each other (Section \ref{sec:wiki_data}). Both setups have advantages and disadvantages. It is more desirable to compare parallel content (setup \#1), but bigger data volumes allow for analyzing lengthier contexts (setup \#2). As mentioned, we preprocessed the data sets with a POS tagger. The POS tagging module used the spaCy \texttt{es\_core\_news\_sm}\footnote{Available at: \texttt{\url{https://spacy.io/models/es}}, June 2024.}  and \texttt{en\_core\_web\_sm}\footnote{Available at: \texttt{\url{https://spacy.io/models/en}}, June 2024.} pipelines for Spanish and English, respectively. In order to balance the categories, we reduced the original seventeen Universal POS tags\footnote{Available at: \texttt{\url{https://universaldependencies.org/u/pos}}, June 2024.} to the following nine: adjectives (ADJ), adpositions (ADP), adverbs (ADV), conjunctions (CONJ), determiners (DET), nouns (NOUN), pronouns (PRON), verbs (VERB), and a last category combining unknown words, interjections, blank spaces, punctuation marks, and symbols (OTHER).

We executed the experiments using two Nvidia A100 GPUs with the specifications in Table \ref{tab:gpu_specs}.

\begin{table}[!t]
\centering
\caption{Specifications of the GPUs.}
\label{tab:gpu_specs}
\begin{tabular}{lc}
\toprule
\multicolumn{2}{c}{\textbf{Nvidia A100-PCIE-40GB specifications}} \\
\midrule
CUDA Driver/Runtime Version & 11.5/11.2\\
CUDA Capability Version & 8.0 \\
Memory & 40 GB (HBM2 bw: 1555 GB/s)\\
Multiprocessors & 108\\
CUDA Cores & 6912 (64 per MP)\\
GPU Max Clock rate: 1.41 GHz\\
\bottomrule
\end{tabular}
\end{table}

\subsection{Tale data sets}
\label{sec:tale_data}

\begin{table}[!htbp]
\centering
\caption{Tale data set sources.}
\label{tab:data_sources}
\begin{tabular}{lc}
\toprule
\textbf{Source} & \textbf{Language} \\
\midrule
Ciudad Seva\tablefootnote{Available at: \texttt{\url{https://ciudadseva.com}}, June 2024.} & Spanish \\
Rincón Castellano\tablefootnote{Available at: \texttt{\url{https://www.rinconcastellano.com}}, June 2024.} & Spanish \\
Elejandría\tablefootnote{Available at: \texttt{\url{https://www.elejandria.com}}, June 2024.} & Spanish \\
Andersenstories.com\tablefootnote{Available at: \texttt{\url{https://www.andersenstories.com}}, June 2024.} & Spanish \& English \\
Grimmstories.com\tablefootnote{Available at: \texttt{\url{https://www.grimmstories.com}}, June 2024.} & Spanish  \& English  \\
Americanliterature.com\tablefootnote{Available at: \texttt{\url{https://americanliterature.com}}, June 2024.} & English \\
D. L. Ashliman's compilation\tablefootnote{Available at: \texttt{\url{https://sites.pitt.edu/~dash/perrault.html}}, June 2024.} & English \\
Long long time ago\tablefootnote{Available at: \texttt{\url{https://www.longlongtimeago.com}}, June 2024.} & English \\
Project Gutenberg\tablefootnote{Available at: \texttt{\url{https://www.gutenberg.org}}, June 2024.} & English \\
The H.P. Lovecraft Archive\tablefootnote{Available at: \texttt{\url{https://www.hplovecraft.com}}, June 2024.} & English \\
\bottomrule
\end{tabular}
\end{table}

The tale data sets of setup \#1 consisted of public domain tales, short novels, and fables with Creative Commons-licensed translations.
We crawled the English and Spanish text collections for the sizes of the respective datasets to be identical (3.7M words of raw text each) so that they could be considered direct translations, 
from Portable Document Format (PDF) with the Python \texttt{pdftotext}\footnote{Available at: \texttt{\url{https://pypi.org/project/pdftotext}}, June 2024.} library and by web scrapping using \texttt{Scrapy}\footnote{Available at: \texttt{\url{https://scrapy.org}}, June 2024.} web spiders. Table \ref{tab:data_sources} shows all the data sources.
 
We could not compute conditional entropy values for contexts longer than two words due to data set size constraints. However, because of the similarities in content between English and Spanish data, we could efficiently study short-term grammatical dependencies in both languages. In the experiment, we compared the causal two-word context with a specific case of non-causal context in which the predicted word lies between the two contextual words. We provide a brief qualitative analysis of the contexts that resulted in lower conditional entropy values for the predicted term $\hat{X}_i$ for causal and bidirectional contexts in both languages.

\begin{table}[!t]
\caption{Two-word context-conditioned entropy in bits per tag, tale data set.\label{tab:tale_pos_n2}}
\centering
\begin{tabular}{l  c  c }\toprule
\textbf{Language}   &  $\olsi{H(\hat{X}_i \mid X_{i-2}, X_{i-1})}$     &  $\olsi{H(\hat{X}_i\mid X_{i-1}, X_{x_i+1})}$ \\
\midrule
Spanish             & 2.3331            & \textbf{1.8444}    \\
English             & \textbf{2.2193}   & 1.9726    \\
\bottomrule
\end{tabular}
\end{table}

Table \ref{tab:tale_pos_n2} shows the results of hypothesis testing on English and Spanish tale data sets. They are coherent with our initial intuition that Spanish is more suited for non-causal text prediction than English. 
However, by examining Table \ref{tab:tale_pos_n2} row-wise, 
middle tag prediction seemed to have lower entropy than causal text prediction in both languages. This does not necessarily mean that non-causal NLG outperformed its causal counterpart, as this experiment disregarded relevant factors, such as initial text generation steps.

\begin{table}[!t]
\caption{Low entropy causal contexts, Spanish tale data set.}
\label{tab:sp_patt_causal}
\centering
\begin{tabular}[t]{lcc}\toprule
$\left(X_{i-2}, X_{i-1}\right)$& $MaxProb\left(\hat{X}_i\right)$ & $H(\hat{X}_i \mid X_{i-2}, X_{i-1})$\\
\midrule
(ADP, DET) & NOUN & 0.8567 \\
(DET, DET) & NOUN & 0.9571 \\
(VERB, DET) & NOUN & 0.9985 \\
(ADV, PRON) & VERB & 1.0101 \\
(PRON, PRON) & VERB & 1.0709 \\
(ADV, DET) & NOUN & 1.0919 \\
(CONJ, DET) & NOUN & 1.1460 \\
(ADJ, DET) & NOUN & 1.2238 \\
(OTHER, DET) & NOUN & 1.3036 \\
(CONJ, PRON) & VERB & 1.3081 \\
(NOUN, PRON) & VERB & 1.4421 \\
(NOUN, DET) & NOUN & 1.4576 \\
(DET, ADJ) & NOUN & 1.5038 \\
\bottomrule
\end{tabular}
\end{table}

\begin{table}[!t]
\caption{Low entropy bidirectional contexts, Spanish tale data set.}
\label{tab:sp_patt_noncausal}
\centering
\begin{tabular}[t]{lcc}\toprule
$\left(X_{i-1}, X_{i+1}\right)$& $MaxProb\left(\hat{X}_i\right)$ & $H(\hat{X}_i \mid X_{i-1}, X_{i+1})$\\
\midrule
(DET, ADP) & NOUN & 0.3316 \\
(DET, OTHER) & NOUN & 0.4059 \\
(DET, CONJ) & NOUN & 0.5770 \\
(ADP, NOUN) & DET & 0.7159 \\
(DET, ADJ) & NOUN & 0.7238 \\
(DET, ADV) & NOUN & 0.8258 \\
(PRON, ADP) & VERB & 0.8354 \\
(DET, PRON) & NOUN & 0.8426 \\
(DET, VERB) & NOUN & 0.9659 \\
(NOUN, CONJ) & OTHER & 1.2998 \\
(PRON, DET) & VERB & 1.3234 \\
(ADJ, CONJ) & OTHER & 1.3343 \\
(PRON, ADV) & VERB & 1.3719 \\
(ADP, OTHER) & NOUN & 1.3813 \\
(VERB, NOUN) & DET & 1.4323 \\
(ADP, CONJ) & NOUN & 1.4736 \\
(CONJ, NOUN) & DET & 1.4812 \\
(PRON, OTHER) & VERB & 1.5257 \\
(ADP, ADP) & NOUN & 1.5785 \\
\bottomrule
\end{tabular}
\end{table}

\begin{table}[!t]
\caption{Low entropy causal contexts, English tale data set.}
\label{tab:en_patt_causal}
\centering
\begin{tabular}[t]{lcc}\toprule
$\left(X_{i-2}, X_{i-1}\right)$& $MaxProb\left(\hat{X}_i\right)$ & $H(\hat{X}_i \mid X_{i-2}, X_{i-1})$\\\midrule
(ADV, PRON)   &	VERB	&   1.0253 \\
(NOUN, DET)	&   NOUN	&   1.0716 \\
(ADP, DET)	&   NOUN	&   1.1451 \\
(CONJ, PRON)  &	VERB	&   1.2328 \\
(OTHER, PRON) &	VERB	&   1.2841 \\
(ADJ, ADJ)    &	NOUN	&   1.2921 \\
(VERB, DET)   &	NOUN	&   1.3121 \\
(DET, ADJ)    &	NOUN	&   1.3140 \\
(CONJ, DET)   &	NOUN	&   1.3340 \\
(OTHER, DET)  &	NOUN	&   1.3978 \\
(ADV, DET)    &	NOUN	&   1.4669 \\
(PRON, DET)   &	NOUN	&   1.4800 \\
(NOUN, NOUN)  &	NOUN	&   1.5256 \\
\bottomrule
\end{tabular}
\end{table}

\begin{table}[!t]
\caption{Low entropy bidirectional contexts, English tale data set.}
\label{tab:en_patt_noncausal}
\centering
\begin{tabular}[t]{lcc}\toprule
$\left(X_{i-1}, X_{i+1}\right)$& $MaxProb\left(\hat{X}_i\right)$ & $H(\hat{X}_i \mid X_{i-1}, X_{i+1})$\\
\midrule
(DET, ADP) & NOUN & 0.3183 \\
(DET, VERB) & NOUN & 0.5168 \\
(DET, ADV) & NOUN & 0.5461 \\
(DET, OTHER) & NOUN & 0.5776 \\
(DET, CONJ) & NOUN & 0.6541 \\
(ADJ, ADP) & NOUN & 0.9530 \\
(ADJ, OTHER) & NOUN & 0.9721 \\
(NOUN, CONJ) & OTHER & 1.0294 \\
(DET, PRON) & NOUN & 1.0911 \\
(ADJ, CONJ) & NOUN & 1.3138 \\
(ADP, ADJ) & DET & 1.3873 \\
(CONJ, VERB) & PRON & 1.4157 \\
(PRON, ADV) & VERB & 1.4580 \\
(NOUN, OTHER) & NOUN & 1.5198 \\
\bottomrule
\end{tabular}
\end{table}

\begin{figure*}[!t]
\centering
\subfloat[]{\includegraphics[width=0.45\textwidth]{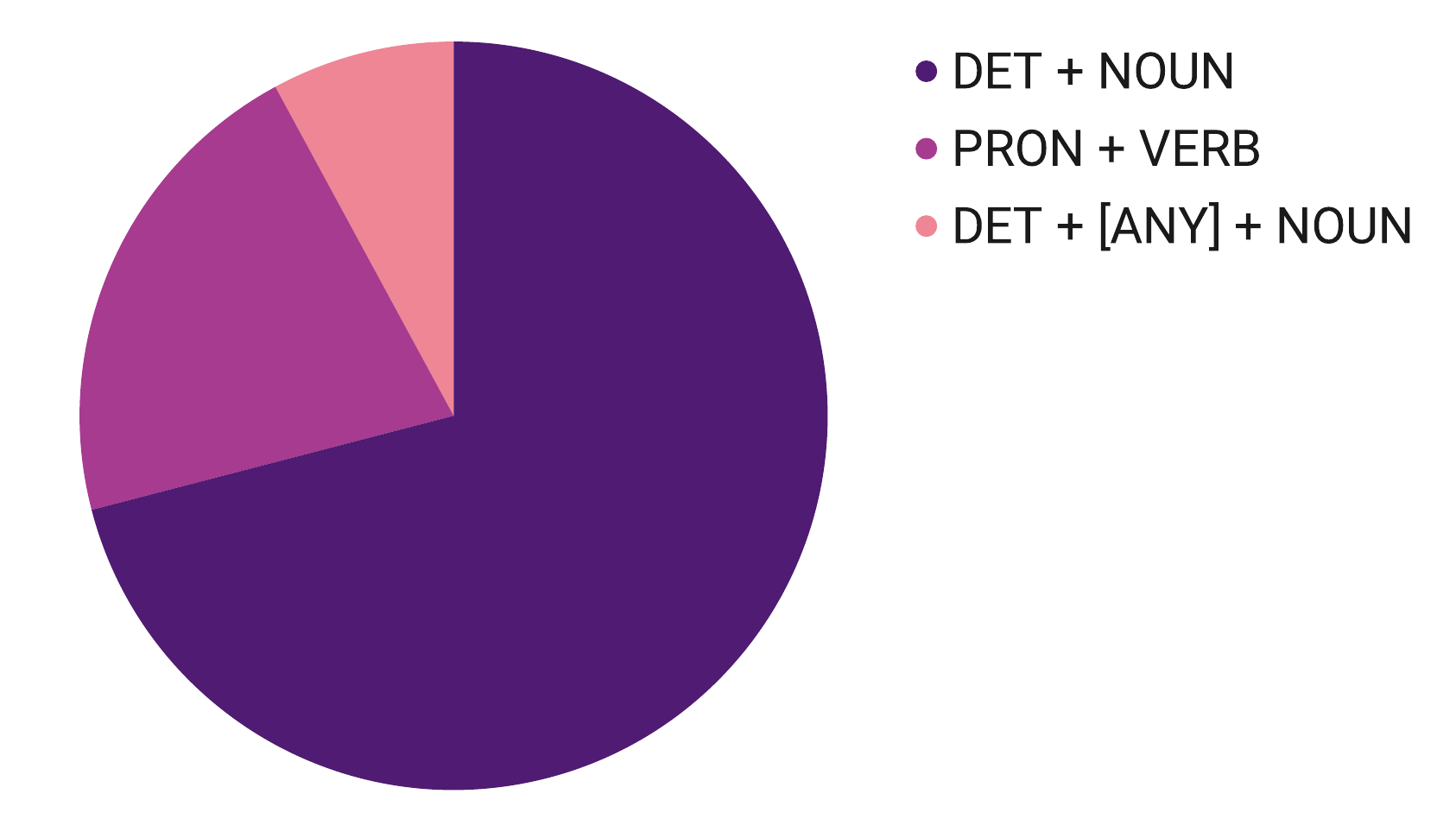}}
\hfil
\subfloat[]{\includegraphics[width=0.45\textwidth]{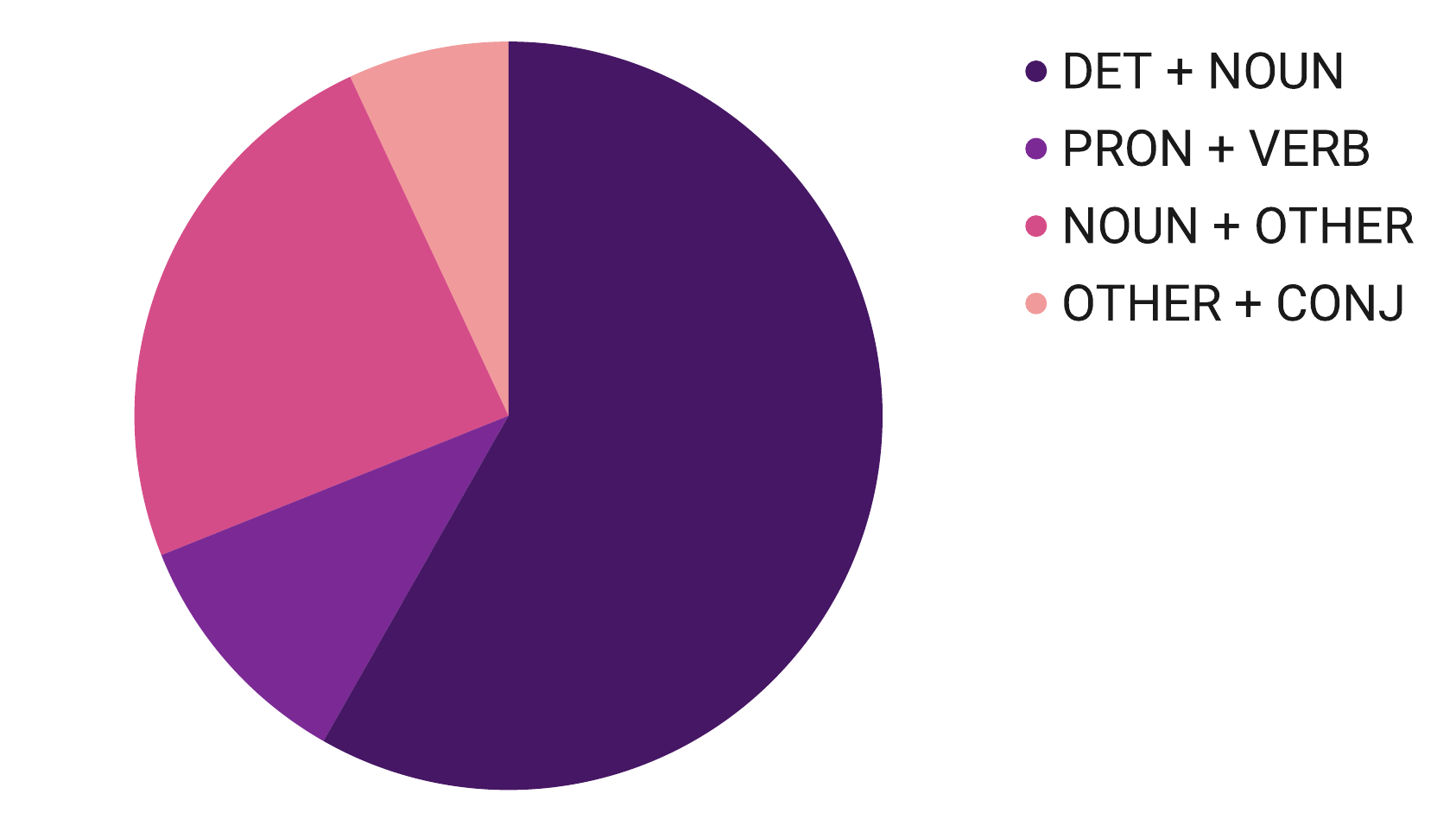}}
\caption{Spanish tale data set low-entropy context pattern distribution. (a) Causal. (b) Bidirectional.}
\label{fig:figure1}
\end{figure*}

\begin{figure*}[!t]
\centering
\subfloat[]{\includegraphics[width=0.45\textwidth]{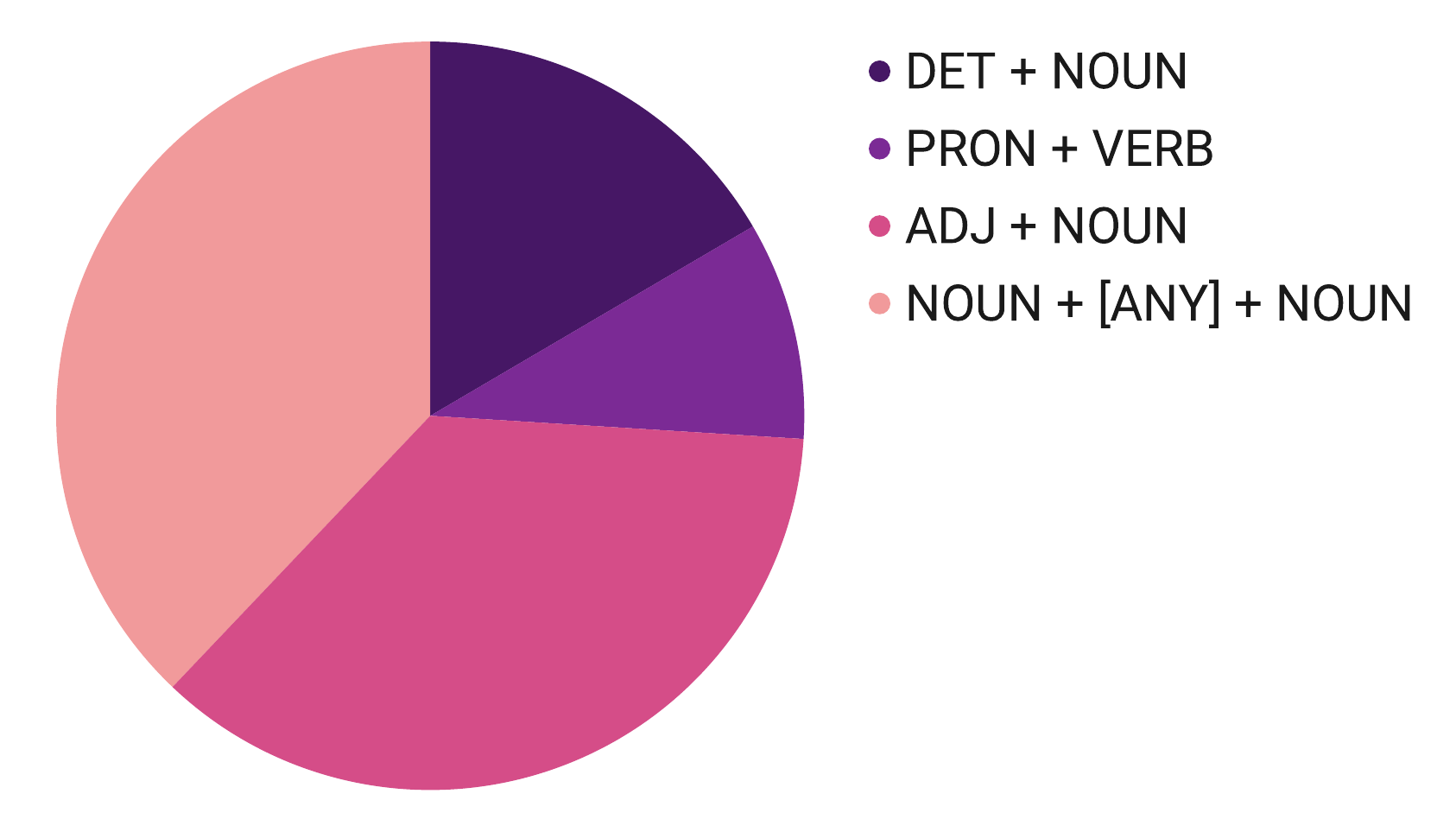}}
\hfil
\subfloat[]{\includegraphics[width=0.45\textwidth]{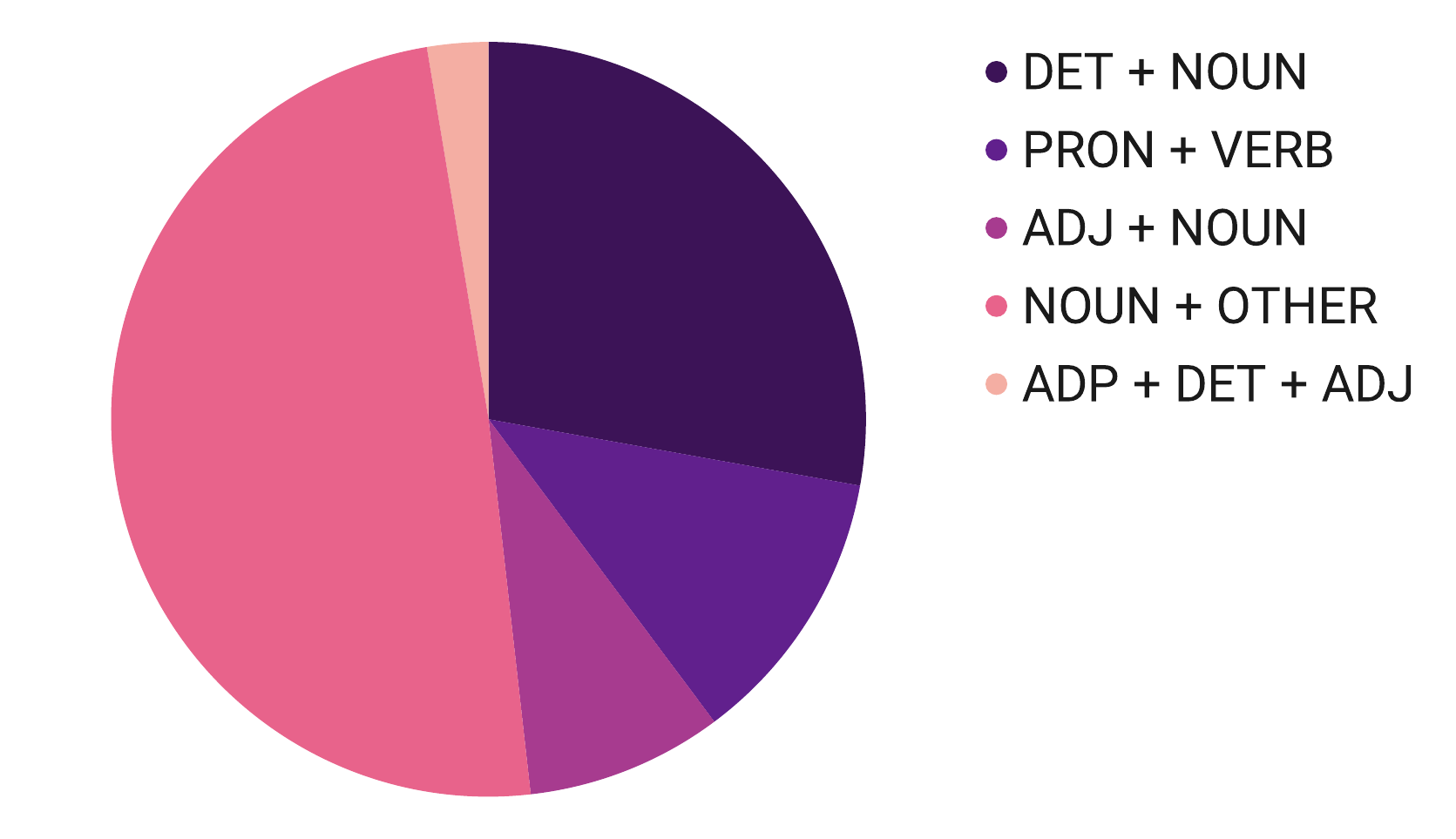}}
\caption{English tale data set low-entropy context pattern distribution. (a) Causal. (b) Bidirectional.}
\label{fig:figure2}
\end{figure*}

Tables \ref{tab:sp_patt_causal}, \ref{tab:sp_patt_noncausal}, \ref{tab:en_patt_causal} and \ref{tab:en_patt_noncausal} show left and bidirectional context-predicted word pairs with entropy lower than $\log(3) \approx 1.585$. We can note that the number of combinations satisfying this condition for the Spanish data set was much higher for bidirectional contexts than for causal contexts. Figure \ref{fig:figure1} shows that, for Spanish, causal patterns were also less diverse, as all of them relied on either pronoun/verb or determinant/noun grammatical dependencies.

As shown in tables \ref{tab:en_patt_causal} and \ref{tab:en_patt_noncausal}, causal and bidirectional low entropy contexts in English were more balanced. Figure \ref{fig:figure2} reflects the prevalence of the adjective/noun dependencies (especially for the causal case). Conversely, adjectives have a much less rigid position in Spanish within the sentence.

\subsection{Wikidumps data sets}
\label{sec:wiki_data}
The data sets of setup \#2 comprised a collection of 
Wikipedia articles from Wikimedia's Spanish\footnote{Available at: \texttt{\url{https://dumps.wikimedia.org/eswiki/20220801}}, June 2024.} and English\footnote{Available at: \texttt{\url{https://dumps.wikimedia.org/enwiki/20220801}}, June 2024.} dumps. We extracted and cleaned textual data from these dumps using \texttt{WikiExtractor}\footnote{Available at: \texttt{\url{https://github.com/attardi/wikiextractor}}, June 2024.}. Then we loaded and mapped the resulting JSON files with HuggingFace's \texttt{Datasets}\footnote{Available at: \texttt{\url{https://github.com/huggingface/datasets}}, June 2024.} library. We picked one million random non-empty articles in each language for hypothesis testing.

The amount of data available allowed accurately computing the conditional entropy for longer contexts than in the previous case, yielding average conditional entropy results for contexts up to six words. For this experiment, we explored all possible non-causal contexts and assessed the impact of the location of the predicted tag on our predictability results.

\begin{table}[!t]
\caption{Conditional entropy in bits per tag, Wikidump data set.\label{tab:wiki_pos}}
\centering
\begin{tabular}{l c  c  c  c}\toprule
\textbf{Context} & \textbf{Language}   &  $\olsi{H(\hat{X}\mid X_{c})}$ &  $\olsi{H(\hat{X}\mid X_{n})}$ & $\olsi{H(\hat{X}\mid X_{n}\setminus X_{c})}$\\
\midrule
\multirow{2}{*}{$N=2$}  & Spanish   & 2.4981  &  2.3192 & \textbf{1.9609} \\
                        & English   & \textbf{2.3335}  &  \textbf{2.2306} & 2.0255 \\
                        \midrule
\multirow{2}{*}{$N=3$}  & Spanish   & 2.4258  &  2.1286 & \textbf{1.8313} \\
                        & English   & \textbf{2.2903}  & \textbf{ 2.1077 }& 1.9257 \\
                        \midrule
                        
\multirow{2}{*}{$N=4$}  & Spanish   & 2.3686  &  \textbf{1.9819} & \textbf{1.7247} \\
                        & English   & \textbf{2.2559}  &  2.0158 & 1.8544 \\
                        \midrule

\multirow{2}{*}{$N=5$}  & Spanish   & 2.3412  &  \textbf{1.8252} & \textbf{1.5679} \\
                        & English   & \textbf{2.2329}  &  1.9399 & 1.7942 \\
                        \midrule
                        
\multirow{2}{*}{$N=6$}  & Spanish   & 2.2817  &  \textbf{1.6351} & \textbf{1.3758} \\
                        & English   & \textbf{2.1846}  &  1.8599 & 1.7200 \\
\bottomrule
\end{tabular}
\end{table}

\begin{figure*}[!t]
\centering
\subfloat[]{\includegraphics[width=0.45\textwidth]{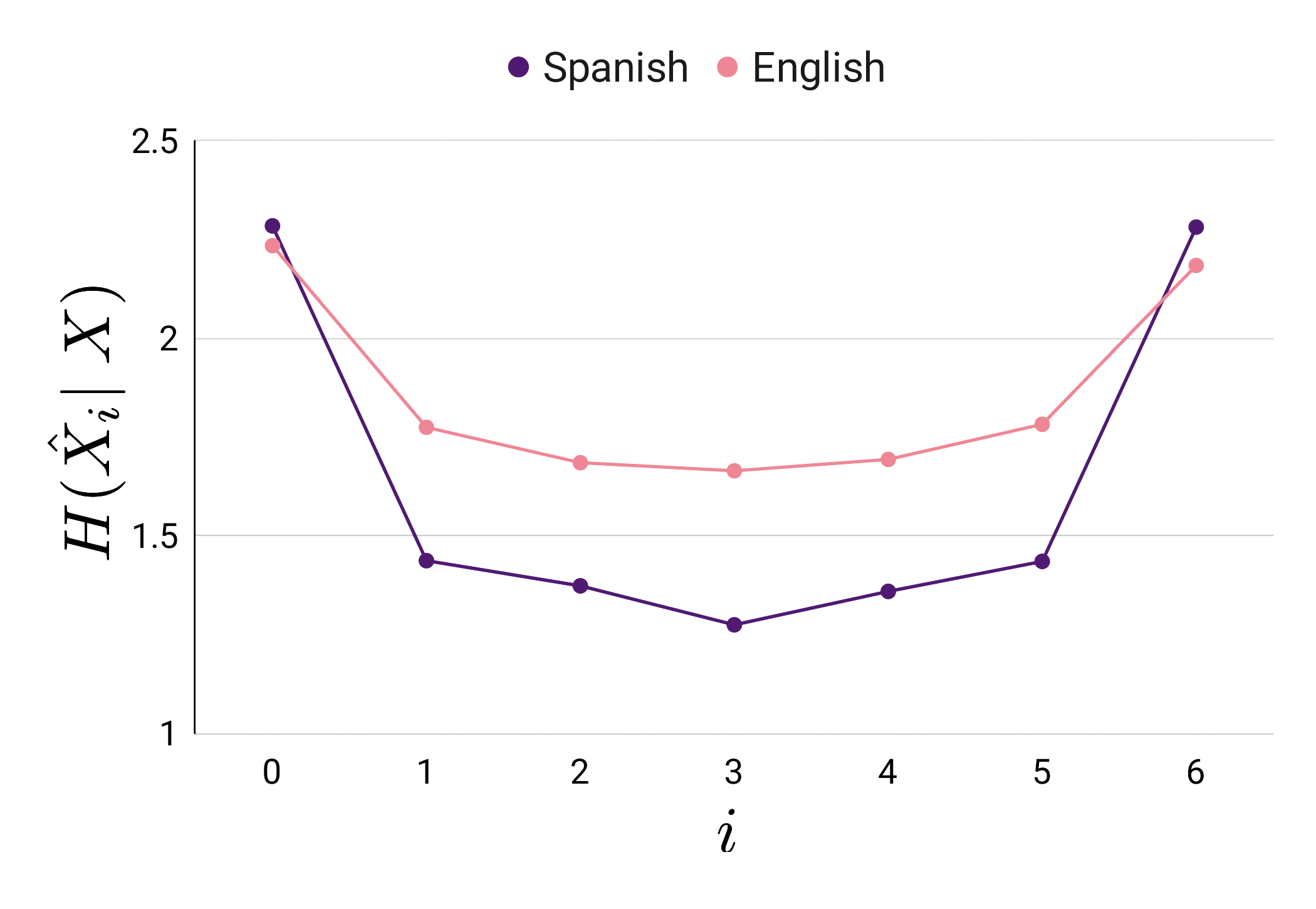}}
\label{fig:3-a}
\hfil
\subfloat[]{\includegraphics[width=0.45\textwidth]{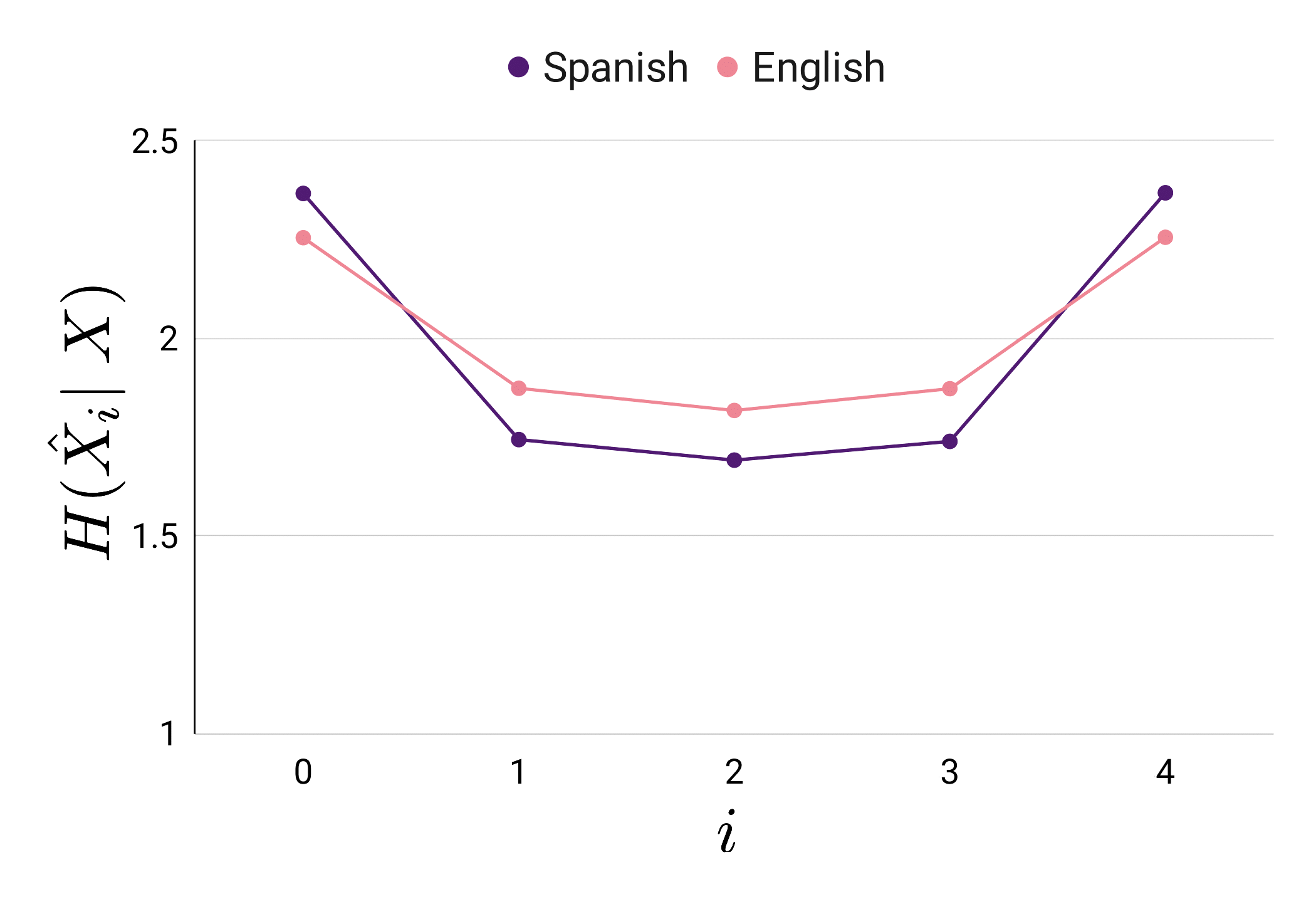}
\label{fig:3-b}}\\
\subfloat[]{\includegraphics[width=0.45\textwidth]{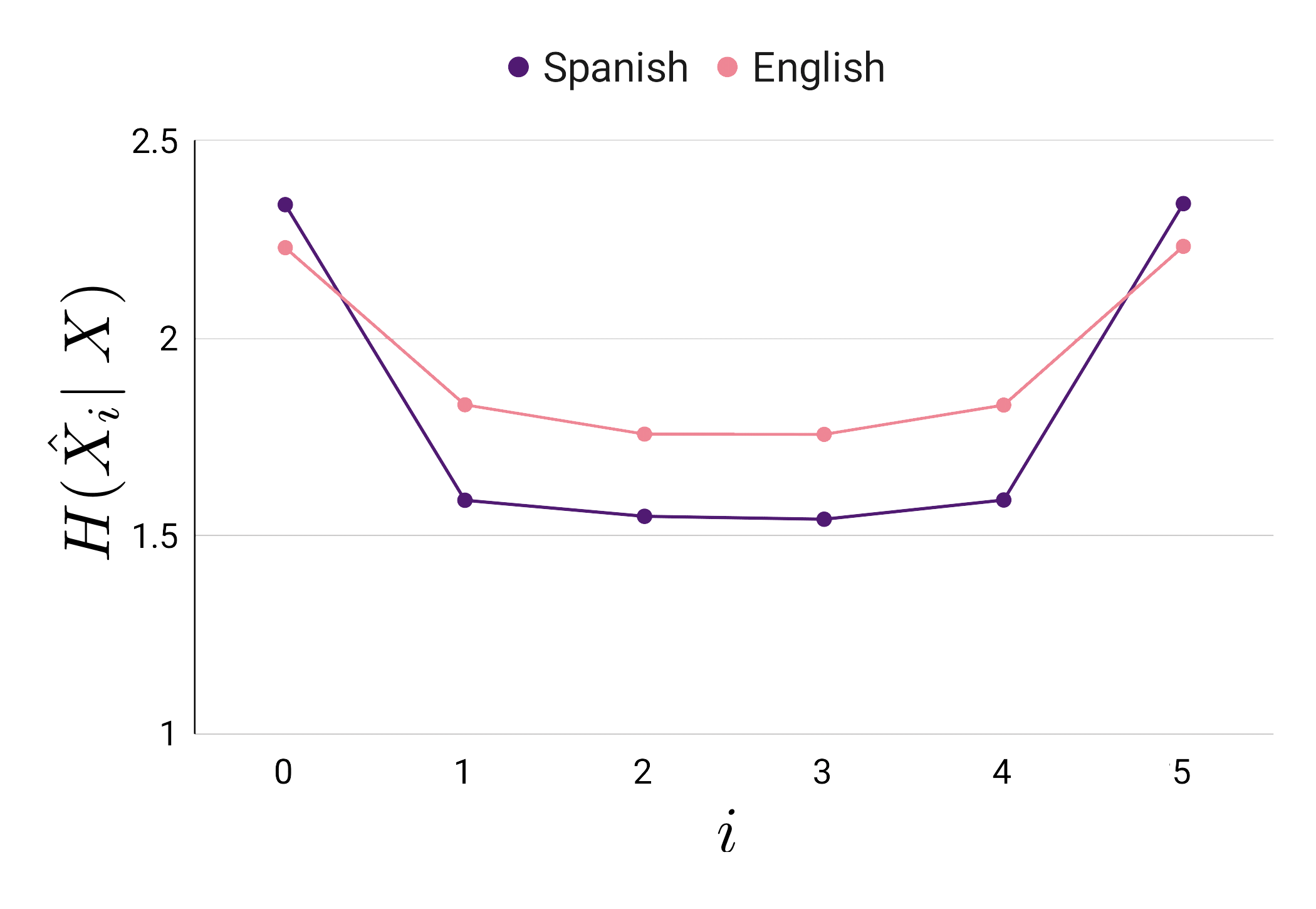}
\label{fig:3-c}}
\hfil
\subfloat[]{\includegraphics[width=0.45\textwidth]{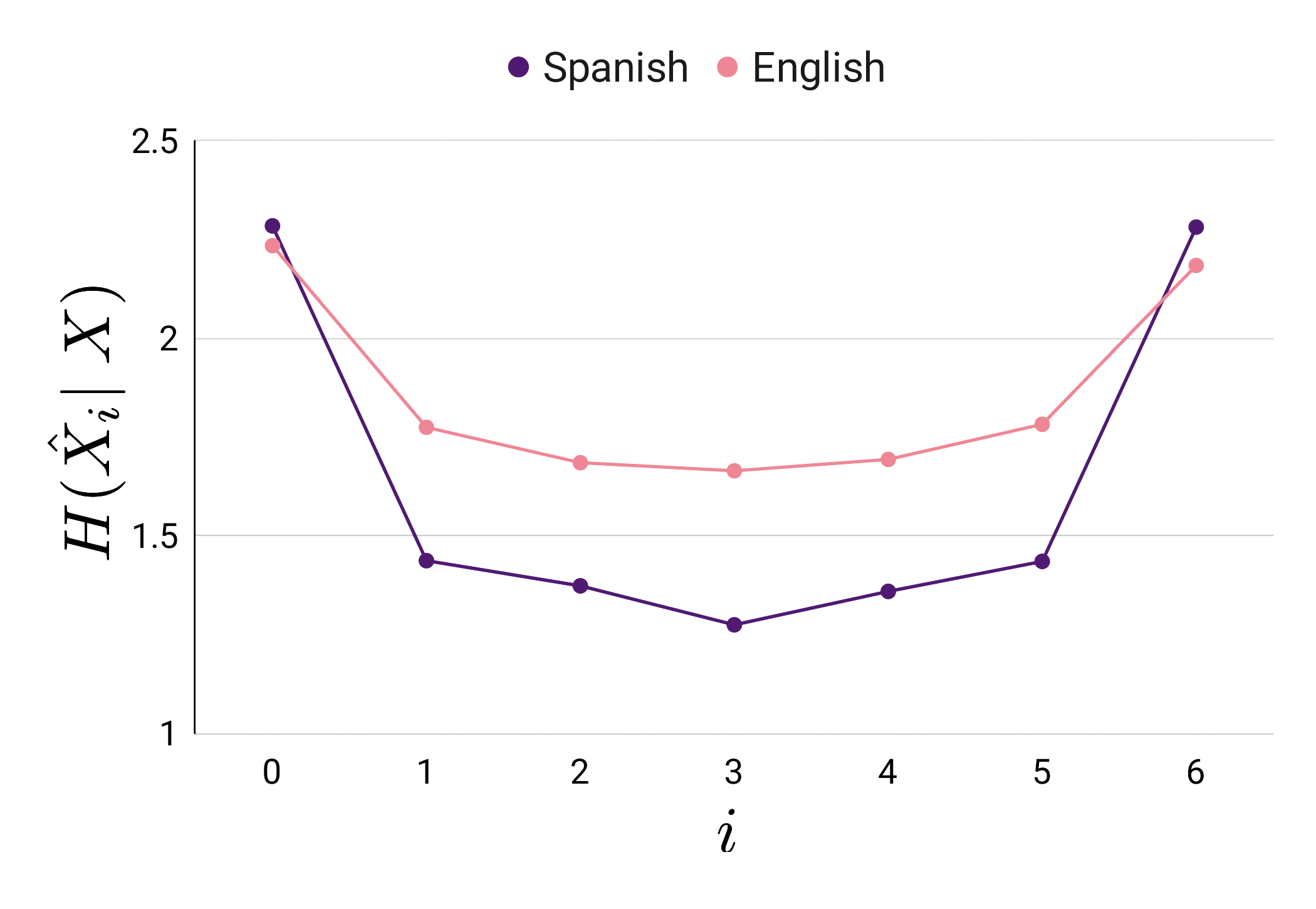}
\label{fig:3-d}}
\caption{Average conditional entropy by predicted tag location for different context lengths. (a) $N=3$. (b) $N=4$. (c) $N=5$. (d) $N=6$.}
\label{fig:figure3}
\end{figure*}

Table \ref{tab:wiki_pos} shows the average causal and non-causal conditional entropy results for the Wikidumps data set. Compared to the tale data set with tighter equivalence between languages, Spanish conditional entropy values often exceeded those for English. For shorter contexts, even non-causal conditional entropy was higher in the Spanish data set than in the English data set. For this reason, we report average entropy as conditioned by the set of non-causal contexts by subtracting the causal context, which is always lower in Spanish than English, highlighting the lower efficiency of left-to-right word prediction in Spanish.

We can further appreciate this effect in Figure \ref{fig:figure3}, which depicts the average conditional entropy for all possible predicted tag locations within contexts of diverse lengths. As we can see, the more balanced the left and right contexts are, the more predictable the grammatical category in both languages. However, this tendency was considerably more noticeable in Spanish and became more evident as the context was longer. Since the average number of words per sentence tends to be higher in Spanish than in English \cite{Bjornsson1983, Harmon1991, Simpson2000}, Spanish non-causal NLG is even more promising.

\section{Text generation results}
\label{sec:text-generation-results}
For the text generation experiment, we used four different language models: Deep ESP's Spanish GPT-2\tablefootnote{Available at: \texttt{\url{ https://huggingface.co/DeepESP/gpt2-spanish}}, June 2024.} and University of Chile's Spanish BERT\tablefootnote{Available at: \texttt{\url{ https://huggingface.co/dccuchile/bert-base-spanish-wwm-cased}}, June 2024.} for Spanish causal and non-causal language modeling, respectively, and OpenAI's GPT-2 small\tablefootnote{Available at: \texttt{\url{ https://huggingface.co/gpt2}}, June 2024.} and Google's BERT base\tablefootnote{Available at: \texttt{\url{ https://huggingface.co/bert-base-cased}}, June 2024.} for English causal and non-causal language modeling, respectively.

Therefore, we used the small versions of these language models, which range from 110M to 117M parameters. We are aware that these models are not on par with state-of-the-art generative language models, but our goal was not to achieve the best text generation results but rather to compare the performance of causal and non-causal language models with similar characteristics. The chosen models were comparable in both training data, number of parameters, and vocabulary size, thus appropriate for this experiment.

We fine-tuned the models on the tales data set described in Section \ref{sec:tale_data} using HuggingFace's Transformers\footnote{{\label{fn:transformers}}Available at: \texttt{\url{https://github.com/huggingface/transformers}}, June 2024.} library. The training was executed in 10 epochs with batch size 8 using a Nvidia A100-PCIE-40GB (see Table \ref{tab:gpu_specs}).

Then, we generated 1,000 different 50-token sequences for each of the four language models. We assessed the generation performance of each language model using both automatic (Section \ref{sec:results_auto}) and manual (Section \ref{sec:results_human}) evaluation metrics.

\subsection{Automatic evaluation}
\label{sec:results_auto}
In the sequel, we will call ``opposite'' the language, either Spanish or English, for which the model was not  trained for an experiment.
We used the conditional relative entropy metric described in Section \ref{sec:auto_metrics} for automatic evaluation to compare the produced sequences from the tales data sets in the target and opposite languages. This was possible because language-independent POS tagging was used for tokenization. 

As expected, Table \ref{tab:auto} reveals that the sequences generated by all four models adhered more closely to their respective target language's POS tag distributions. There were no significant differences between causal and non-causal context-conditioned relative entropy results. 

Given its conditional relative entropy values one order of magnitude greater than those of the other three models,
English BERT performed significantly worse in terms of 
adherence to the target language ($\sim$0.02 vs $\sim$0.1),
which is consistent with the state of the art \cite{Wang2019a, Shen2020}.
For Spanish, the results of causal and non-causal NLG are more comparable. The Spanish BERT non-causal  language model had the lowest conditional relative entropy, outperforming even English GPT-2 when considering adherence to the respective target languages. 

\begin{table}[hbt]
\centering
\caption{Conditional relative entropy results.``Causal'' and ``non-causal''  refer to causal and non-causal context-conditioned relative entropy, respectively.}
\begin{tabular}{lcccc}\toprule
\multicolumn{1}{c}{} & \multicolumn{2}{c}{\bf Spanish Data Set} & \multicolumn{2}{c}{\bf English Data Set}\\
      & causal & non-causal & causal & non-causal \\
\textbf{Model} &  &  &  &  \\
\midrule
Spanish GPT-2  & 0.0351  & 0.0387  & 1.1903 & 1.2588\\
Spanish BERT & \textbf{0.0210}  & \textbf{0.0221}  & 1.3512 & 1.4049\\
\midrule
English GPT-2 & 1.0096  & 1.5505  & \textbf{0.0245} & \textbf{0.0292}\\
English BERT  & 0.8817  & 1.4437  & 0.1010  & 0.1050\\
\bottomrule
\end{tabular}
\label{tab:auto}
\end{table}

These results show that causal models corresponded more closely to English grammar and non-causal models corresponded more closely to Spanish grammar (by considering grammar as reflected in the English and Spanish datasets). 
Note that this also held when using the models of the opposite language. That is, the conditional relative entropy of English Bert for Spanish was lower than the conditional relative entropy of English GPT-2 for Spanish, and the conditional relative entropy of Spanish GPT-2 for English  was lower than the conditional relative entropy of Spanish BERT for English. 

\subsection{Manual evaluation}
\label{sec:results_human}

We chose 250 sequences at random for each pairing of language (Spanish or English) and model (BERT or GPT-2)  to reduce the annotation load while still obtaining useful insights. Each sequence was examined independently using the questions in Section \ref{sec:human_method}.

Five annotators participated in the evaluation. Table \ref{tab:i-a_global} shows the global inter-agreement analysis of yes/no replies. Using the thresholds by \cite{Landis1977}, $ \alpha$-reliability coefficients lay between fair and moderate agreement. Accuracy values were also acceptable for all questions.
The grammatical structure (Q2)  was the most controversial aspect of the first four questions due to different opinions on linguistic demand, as some annotators were more lenient with one of the languages.

\begin{table}[hbt]\centering
\caption{Global inter-agreement metrics for yes/no questions.}\label{tab:i-a_global}

\begin{tabular}{l c c c c}\toprule
& \textbf{Q1} & \textbf{Q2} & \textbf{Q3} & \textbf{Q4}\\
\midrule
Accuracy             & 0.906 & 0.818 & 0.829 & 0.966 \\
$\alpha$-reliability & 0.296 & 0.242 & 0.442 & 0.491 \\
\bottomrule
\end{tabular}
\end{table}

\begin{table}[hbt]\centering
\caption{Manual evaluation results for yes/no questions.}\label{tab:manual_eval}

\begin{tabular}{lcccc}\toprule
\multicolumn{1}{c}{} & \bf Q1 & \bf Q2 & \bf Q3 & \bf Q4\\

\textbf{Model} & \multicolumn{1}{c}{(`yes' \% )} & \multicolumn{1}{c}{(`yes' \% )} & \multicolumn{1}{c}{(`no' \% )} &\multicolumn{1}{c}{(`yes' \% )}\\
\midrule
Spanish GPT-2  & 91.6  & 90.8  & 77.6 & 96.4 \\
Spanish BERT & \textbf{95.2}  & \textbf{91.6}  & \textbf{85.6} & \textbf{98.0} \\
\midrule
English GPT-2  & \textbf{99.2}  & \textbf{93.6}  & \textbf{89.2}  & \textbf{98.4}\\
English BERT & 96.4  & 89.2  & 84.8 & 94.8 \\
\bottomrule
\end{tabular}
\end{table}

Next, we can see that the manual assessment scores for all four questions in Table \ref{tab:manual_eval} were consistent with the automatic metrics: BERT was considered to perform better in Spanish and GPT-2 to perform better in English. The best outcome for Spanish language models was in word sense (Q4), whereas English language models scored better in word concordance (Q1). This is consistent with the fact that there is no gender concordance in English. For all language models, the more challenging question was word repetition (Q3).

\begin{table}[hbt]\centering
\caption{Manual evaluation results, general assessment question.}\label{tab:manual_q5}

\begin{tabular}{lcc}\toprule
\textbf{Model} & \bf Average & \bf Norm. Average\\\midrule
Spanish GPT-2  & 3.682  & 0.920 \\
Spanish BERT & \textbf{4.124} & \textbf{1.034}\\
\midrule
English GPT-2  & \textbf{4.301} & \textbf{1.078}\\
English BERT & 3.870 & 0.967\\
\bottomrule
\end{tabular}
\end{table}

For the more subjective fifth question, as the average rating from one annotator to another ranged from $3.529$ to $4.31$, we normalized the scores of each annotator for an average of 1.0.  The results of Table \ref{tab:manual_q5} indicate that the subjectively perceived quality of Spanish texts generated by BERT is higher than when using GPT-2, vice versa in the case of English, which is consistent with our initial intuition and all the results so far. 

\section{Discussion}
\label{sec:discussion}
All the results of the previous section are aligned with our initial intuition that Spanish is more suited for non-causal language modeling than English.
As shown in Table \ref{tab:summary_pred}, natural English text was demonstrated to be more predictable than text in Spanish given a causal context in the predictability test results, by a relatively constant margin of $\sim$5\%. However, given a non-causal context, Spanish was more predictable than English, by an increasing margin as the context got longer.

In the automatic evaluation with conditional relative entropy, Spanish BERT showed the highest adherence to its target language grammar.

These results are consistent with the text generation ranking summarized in Table \ref{tab:summary_gen} (whose row ``automatic evaluation'' reflects the conditional relative entropies in Table \ref{tab:auto}), as English GPT-2 performed better than English BERT, and Spanish BERT better than Spanish GPT-2 in all the evaluation experiments. Manual evaluation consistently ranked English GPT-2 and Spanish BERT as the best language models for NLG. Spanish BERT ranked  worse than English BERT in concordance, but, as previously stated, this results might be biased by the lack of gender concordance in English.

Overall, the results of our experiments show that non-causal language modeling is more promising for Spanish NLG than for English.

\begin{table}[hbt]\centering
\caption{Predictability test, automatic evaluation results summary}\label{tab:summary_pred}
\begin{tabular}{lcccc}\toprule
\multicolumn{1}{c}{} & \multicolumn{1}{c}{} & \multicolumn{2}{c}{\bf Highest predictability}\\
\textbf{Data set} & \textbf{Context length} &$\olsi{H(\hat{X}\mid X_{c})}^{-1}$ & $\olsi{H(\hat{X}\mid X_{n}\setminus X_{c})}^{-1}$\\
\midrule
Tales & $N=2$ & English ($+5.13\%$)& Spanish($+6.95\%$)\\
\midrule
\multirow{5}{*}{Wikidumps} & $N=2$ & English ($+7.05\%$) & Spanish ($+3.29\%$)\\
& $N=3$ & English ($+5.92\%$)& Spanish ($+5.15\%$)\\
& $N=4$ & English ($+4.99\%$)& Spanish ($+7.52\%$)\\
& $N=5$ & English ($+4.85\%$)& Spanish ($+14.43\%$)\\
& $N=6$ & English ($+4.45\%$)& Spanish ($+25.02\%$)\\
\bottomrule
\end{tabular}
\end{table}

\begin{table}[hbt]\centering
\caption{Text generation ranking, manual evaluation results summary}\label{tab:summary_gen}
\begin{tabular}{lcccc}\toprule
& \multicolumn{2}{c}{\bf Spanish} & \multicolumn{2}{c}{\bf English}\\
\textbf{Experiment} &  GPT-2 & BERT & GPT-2 & BERT\\
\midrule
Automatic evaluation &  \#3 & \#1 & \#2 & \#4 \\
\midrule
Q1. Concordance & \#4 & \#3 & \#1 & \#2 \\
Q2. Syntactic structure & \#3 & \#2 & \#1 & \#4 \\
Q3. Repetitions & \#4 & \#2 & \#1 & \#3 \\
Q4. Word sense & \#3 & \#2 & \#1 & \#4 \\
Q5. General rating & \#4 & \#2 & \#1 & \#3 \\
\bottomrule
\end{tabular}
\end{table}

\section{Conclusions}
\label{sec:conclusion}
In this paper, we have first assessed English and Spanish predictability given causal and non-causal contexts, demonstrating that Spanish is more predictable given a non-causal context.For this purpose, we developed and computed a novel metric of the average causal and non-causal context-conditioned entropies of the grammatical categories present in similar and strictly parallel English and Spanish textual data sets. The experiments have shown that average causal context-conditioned entropy is higher in Spanish texts than in English texts, and that average non-causal context-conditioned entropy is higher in English texts than in Spanish ones. This was further supported by a a study of the grammatical dependencies that are more predictable in each language and how word location within a context influences predictability.

Following the validation of the hypothesis about the relation between causal- and non-causal contexts and language predictability,  we selected causal and non-causal language generators based in Spanish and English models to analytically  assess their quality depending on the target language to generate. To make experiments comparable, we chose similarly dimensioned unidirectional and bidirectional pre-trained transformer language models and fine-tuned them using highly equivalent Spanish and English data sets.

Finally, we evaluated the outcome  both analytically  and manually  to assess the performance of  text generation in all test scenarios. In the first case,  to compare the compliance of the language models with the grammatical structure of their target languages, we have proposed a conditional relative entropy metric. Manual evaluation, which was validated using inter-agreement metrics, 
is coherent with the automatic evaluation, validating it.

The insights of this study  suggest the interest of further research into analyses of language predictability in languages other than English, as well as on efficient text production using bidirectional transformers in Spanish and other languages with similar grammatical structures.

\bibliography{bibliography}{}
\bibliographystyle{IEEEtran}

\begin{IEEEbiography}[{\includegraphics[width=1in,height=1.25in,clip,keepaspectratio]{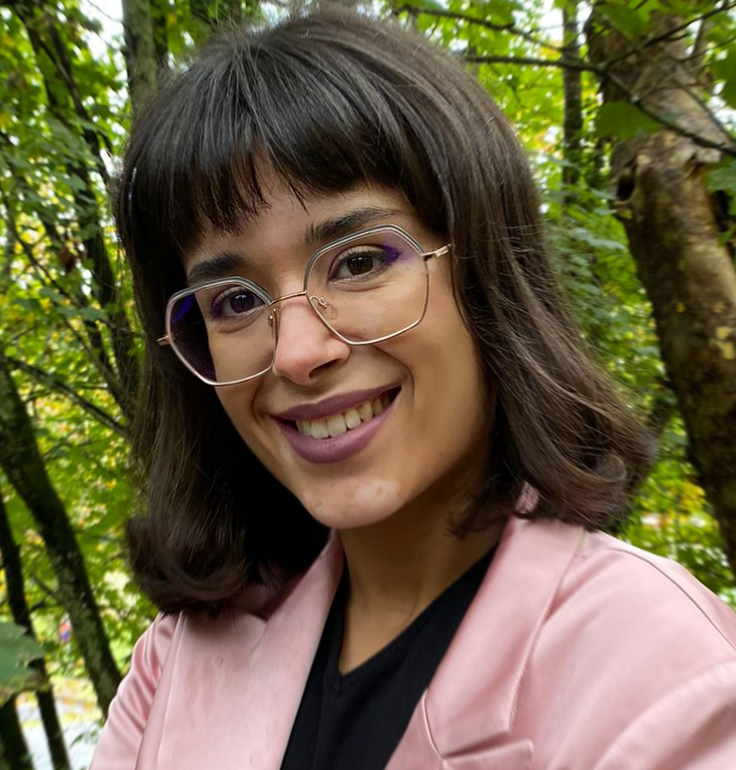}}]{Andrea Busto-Castiñeira} received the B.S. degree in Telecommunication Technologies Engineering in 2020, and the M.S. degree in Telecommunication Engineering in 2022 from University of Vigo, Spain, where she is currently pursuing the Ph.D. degree in Information and Communication Technologies. Since 2021. she has been working as a researcher with the Information Technologies Group. Her research interests include Natural Language Processing and pre-trained language models.
\end{IEEEbiography}

\begin{IEEEbiography}[{\includegraphics[width=1in,height=1.25in,clip,keepaspectratio]{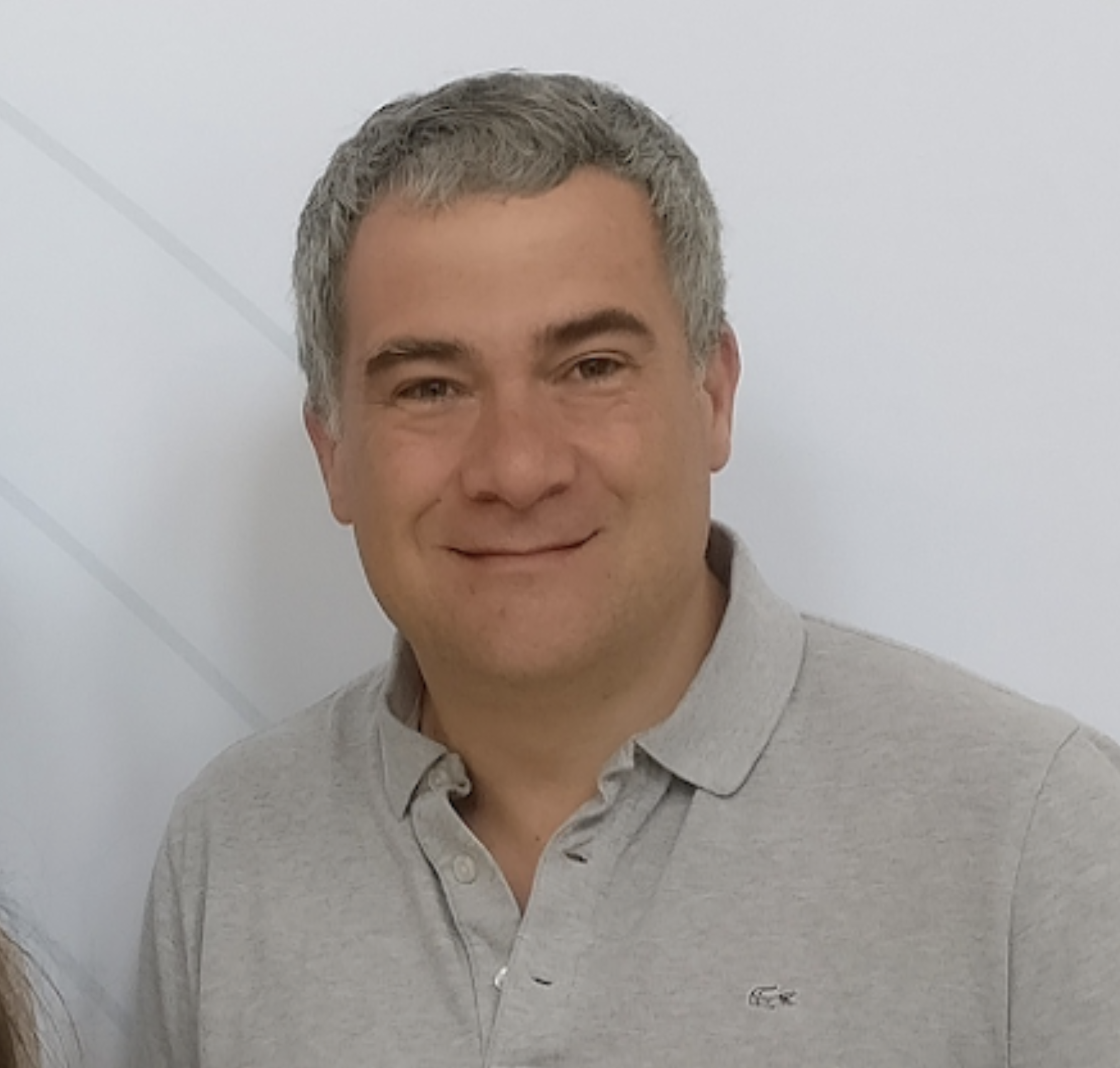}}]{F. Javier González-Castaño} received the B.S. degree from University of Santiago de Compostela, Spain, in 1990, and the Ph.D. degree from University of Vigo, Spain, in 1998. He is currently a professor at University of Vigo, Spain, where he leads the Information Technologies Group. He has authored over 100 papers in international journals in the ﬁelds of telecommunications and computer science, and has participated in several relevant national and international projects. He holds three U.S. patents.
\end{IEEEbiography}

\begin{IEEEbiography}[{\includegraphics[width=1in,height=1.25in,clip,keepaspectratio]{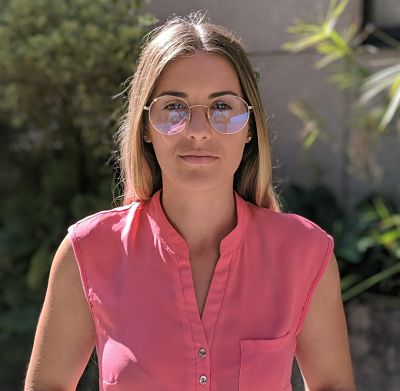}}]{Silvia García-Méndez} received the Ph.D. degree in Information and Communication Technologies from University of Vigo in 2021. Since 2015, she has been working as a researcher with the Information Technologies Group at University of Vigo. She is currently collaborating with foreign research centers as part of her postdoctoral stage. Her research interests include Natural Language Processing techniques and Machine Learning algorithms.
\end{IEEEbiography}

\begin{IEEEbiography}[{\includegraphics[width=1in,height=1.25in,clip,keepaspectratio]{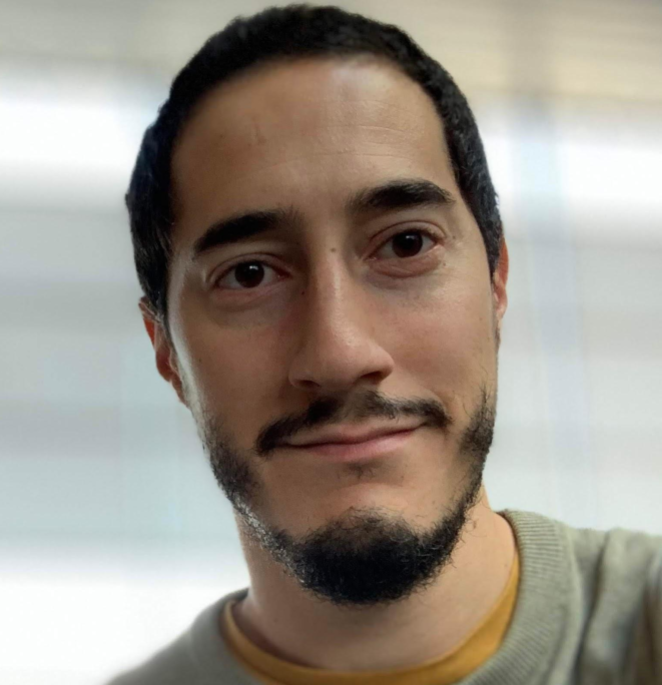}}]{Francisco de Arriba-Pérez} received the B.S. degree in telecommunication technologies engineering in 2013, the M.S. degree in telecommunication engineering in 2014, and the Ph.D. degree in 2019 from University of Vigo, Spain. He is currently a researcher in the Information Technologies Group at the University of Vigo, Spain. His research includes the development of Machine Learning solutions for different domains like finance and health.
\end{IEEEbiography}
\EOD

\end{document}